%% file: cvprw.tex
\documentclass[final]{cvpr}

\usepackage[pdftex]{graphicx}

\usepackage{times}
\usepackage{epsfig}
\usepackage{graphicx}
\usepackage{amsmath}
\usepackage{amssymb}

\usepackage{multirow}

\usepackage[pagebackref=true,breaklinks=true,colorlinks,bookmarks=false]{hyperref}

\def\cvprPaperID{6} 

\def\cvprPaperID{6} 
\def\confYear{CVPR 2021}

\begin{document}

\pagestyle{empty} 

\thispagestyle{empty}

\title{EQFace: A Simple Explicit Quality Network for Face Recognition}

\author{Rushuai Liu\\
Shenzhen Deepcam Information Technologies\\
Shenzhen, China\\
{\tt\small rushuai.liu@deepcam.com}
\and
Weijun Tan\\
LinkSprite Technologies\\
Longmont, CO, USA\\
{\tt\small weijun.tan@linksprite.com}
}

\maketitle
\thispagestyle{empty}

\begin{abstract}
\input{content/abstract.tex}
\end{abstract}

\section{Introduction}
   \input{content/introduction}

\section{Related work}
   \input{content/relate_work}

\section{Proposed Explicit Quality Network}
   \input{content/face_quality.tex}

%

\section{Experiments}
   \input{content/experiment.tex}

\section{Ablation Study and Applications}

\input{content/feature_fuse.tex}

\section{Conclusions}
   \input{content/conclusion.tex}

{\small
\bibliographystyle{splncs}
\bibliography{cvprw}
}

\end{document}

%% file: content/abstract.tex
As the deep learning makes big progresses in still-image face recognition, unconstrained video face recognition is still a challenging task due to low quality face images caused by pose, blur, occlusion, illumination etc. In this paper we propose a network for face recognition which gives an explicit and quantitative quality score at the same time when a feature vector is extracted. To our knowledge this is the first network that implements these two functions in one network online. This network is very simple by adding a quality network branch to the baseline network of face recognition. It does not require training datasets with annotated face quality labels. We evaluate this network on both still-image face datasets and video face datasets and achieve the state-of-the-art performance in many cases. This network enables a lot of applications where an explicit face quality scpre is used. We demonstrate three applications of the explicit face quality, one of which is a progressive feature aggregation scheme in online video face recognition. We design an experiment to prove the benefits of using the face quality in this application. Code will be available at \url{https://github.com/deepcam-cn/facequality}.   

%% file: content/introduction.tex
While face recognition makes tremendous progresses in recent year thanks to deep learning particularly deep CNN, noises in face images, or low-quality face image is still a challenging problem.  For low quality face images caused by pose, blur, occlusion, illumination etc., even though human’s eyes cannot recognize them, a feature vector is still generated, and very likely ends up with a recognition error.  In addition, in the training of a face recognition model, low quality face images are more like noise \cite{WangNoise} and degrade the face recognition performance if not handled appropriately.  For such hard cases, human usually gives an uncertain answer.  Likewise, it is good to have a similar confidence or quality score of uncertainty for every extracted feature vector and use it in training or other scenarios. 

In the video face recognition, this brings further values. In video surveillance applications, the system does not care if every face image in the video is detected and recognized. What really matters is if a person can be recognized correctly during the time he shows up in the video. This can be accomplished by picking up good quality face images, if there are any. This goal is consistent with the bench-marking protocol of the video face data sets, such as the Youtube Face data sets (YTF) \cite{YTF} and the IJB face data sets \cite{IJBA}, \cite{IJBB},\cite{IJBC}, where the verification or identification accuracy is measured per video instead of per image.   

Our contribution in this paper is that we propose a simple explicit quality network for face recognition, which we call EQFace.  This network has the following features.  

1) This network generates an explicit quantitative quality score for a face image. To our knowledge this is the first network that achieves this goal while extracting the feature vector at the same time.  With this explicit quality score available, more ways to use the face quality can be further explored.   

2) This network does not require face datasets with annotated quality value, which is very hard to get.

3) This network is very simple in terms of extra complexity on top of a baseline face recognition network. We add a quality assessment branch network to generate the quality score. A new loss function and a new training pipeline are proposed to work on this branch network.

After the explicit quantitative quality score becomes available in the EQFace, we use it to weight or filter out low quality face images in training and/or testing the face recognition network.  We evaluate the performance on the still-image face datasets LFW \cite{LFWTech}, CALFW \cite{CALFW}, CPLFW \cite{CPLFWTech}, and CFPFP \cite{cfp-paper}, and the template video datasets Youtube face (YTF) \cite{YTF} and IJB-B \cite{IJBB}, IJB-C \cite{IJBC}.  Experimental results confirm the performance improvement brought by weighting or filtering out the feature vector of low qualities in training and/or testing.      

Our second contribution is that we propose a progressive feature aggregation using the explicit face quality out of the EQFace for online video face recognition. Unlike in the bench-marking protocol of YTF and IJB data sets, where feature aggregation is done after all video frames are processed, we aggregate the features online while video frames become available. we design an experiment to evaluate the performance of this  feature aggregation scheme. Our results show impressive performance improvement.

%% file: content/relate_work.tex
Quite a few approaches have been studied to combat the problem of using noisy data to learn the face recognition representation, including \cite{Noise1},\cite{Noise2},\cite{Noise3},\cite{Noise4},\cite{Noise5}.  In \cite{WangNoise}, a co-mining strategy using the loss value as the cue of noisy label was proposed. However, the very complex twin network made the training on a large dataset very challenging.  Our idea is along the line of reweighting, but weighted by an explicit and quantitative face quality.  As pointed out in \cite{Subcenter}, re-weighting methods are susceptible to the performance of the initial model. We work around this problem by proposing an iterative training pipeline.  

An earlier work that gave explicit and quantitative face quality is \cite{faceQualityPrediction}. It is a good reference for review of quality assessment work before 2017. They studied the face quality assessment using a SVR predictor on extracted CNN features. In order to do so, they collected training dataset with quality label annotated by human. This is not only very time consuming, but also subjective to the annotating person. They compared the quality assessments by human and by the SVM predictor. They then used the quality generated by the SVR predictor to filter out low quality faces in face recognition. 

Another work that gave quantitative face quality is the FaceQNet \cite{FaceQnet}. It proposed to find the best quality face image for every face ID, and use it as the base for its class.  After that, the face recognition model is used to calculate the similarity scores between this face and all other faces. And these scores are used as quality labels for these faces. This approach has some disadvantages.  For example, one person has two face images taken at different ages, the qualities of both images are good, but their cosine similarity may not be large, therefore may not be appropriate to be used as the quality score. 

The authors of \cite{CrystalLoss} studied the benefits from using the quantitative face quality in the face recognition network. They also studies a feature aggregation scheme by using the face quality. However, they did not study how to obtain the face quality. Instead, they simply used the confidence score of face detection as the quality score. In our study, we found that the face detection score is not a good metric for face quality. 

QAN \cite{QAN} is the first work that explored a face quality prediction function embedded in the baseline network. In their work, a two-branch network architecture is used. The first branch extracts feature vector for every sample of data, while the second branch predicts a quality score for the sample.  Then the features and quality scores of all samples in a set are aggregated to generate the final feature vector.  However, QAN did not give an explicit face quality output.  

Inspired by QAN, the NAN \cite{NAN} and \cite{FANVFR} were proposed.  The NAN used a two layers of neurons to train the quality weight in the feature aggregation. The \cite{FANVFR}  extended NAN by replacing the per frame scalar weight with a matrix weight.  Both did not generate an explicit face quality score.  

In \cite{AttentionVFP} an attention-aware method was proposed for video face recognition. They modeled the attentions of videos as a Markov decision process and train the attention model through a deep reinforcement learning. The motivation was to discard the misleading and confusing frames and find the focuses of attentions in face videos. They used the information in both the image space and the feature space. Same as other work, it did not give an explicit face quality score. 

Our work is inspired mostly by the work in \cite{conf/iccv/ShiJ19} and \cite{journals/corr/abs-2002-11841}, particularly by \cite{journals/corr/abs-2002-11841}, where they first introduced quality weighing in their loss functions.  We leave more detailed discussion of their work in next section.

%% file: content/face_quality.tex
In this section we describe a new approach - the so called EQFace to estimate the quality of face image. Other than \cite{conf/iccv/ShiJ19} and \cite{journals/corr/abs-2002-11841}, our work is inspired by the ArcFace \cite{ArcFace} for their superior performance in face recognition. 

\subsection{Review of ArcFace Loss Functions}

In the training of a face recognition model,in order to decrease the distance inside a class, and increase the distance between classes, many loss functions have been proposed. Among them, SphereFace \cite{SphereFace} CosFace\cite{CosFace}, ArcFace\cite{ArcFace} added margin into different factor to achieve better training results. The CircleLoss\cite{sun2020circle} unified the interpretation of the loss function from another view, and proposed more reasonable weight assignments to the probabilities of the positive and negative samples. These methods helped significantly to improve the face recognition performance.

In \cite{ArcFace}, starting with the widely used softmax loss, the authors proposed the ArcFace: additive angular margin loss for face Recognition. Let $f_i \in R^d $ denote the feature vector of the sample $x_i$, belonging to the $y_i$-th class. The feature vector dimension $d$ is typically 512. Let $W_j \in R^d$ denote the $j-$th column of the weight $W \in R^{d \times n}$ and $b_j \in R^n$ denote the bias term. Let the class number be $n$. The softmax loss is expressed as: 

\begin{equation}
    l_1 = -log\frac{e^{W_{y_i}^Tf_i+b_{y_i}}}{\sum_{j=1}^n e^{W_{j}^Tf_i+b_{j}}}
    \label{l1}
\end{equation}

Please note that we do not include the summation over batch size in all loss functions in this paper for simplicity. The bias term can usually be fixed $b_j = 0$. Then the logit can be transformed to $W^T_j f_i = ||W_j| |||x_i|| cos(\theta_j)$, where $\theta_j$ is the angle between the weight $W_j$ and the feature $x_i$. As usual, the individual weight can be fixed to $||W_j|| = 1$ by L2 normalisation. Similarly, the feature vector $f_i$ is fixed to $||f_i||=1$ by L2 normalisation and then re-scale it to $s$. After adding a margin term, the ArcLoss is presented as:   

\begin{equation}
    l_2 = -log\frac{e^{s(cos(\theta_{y_i}+m))}}{e^{s(cos(\theta_{y_i}+m))}+\sum_{j\ne{y_i}} e^{s(cos(\theta_{j}))}}
    \label{l2}
\end{equation}

By combining all of the margin penalties, they implemented the SphereFace, ArcFace and CosFace in an united framework with $m1$, $m2$ and $m3$ as the hyper-parameters:

\begin{equation}
    l_3 = -log\frac{e^{s(m_{1}cos(\theta_{y_i}+m_{2})-m3)}}{e^{s(m_{1}cos(\theta_{y_i}+m_{2})-m3)}+\sum_{j\ne{y_i}} e^{s(cos(\theta_{j}))}}
    \label{l3}
\end{equation}

Our loss function is based on the loss function in Equ. (\ref{l3}) for its superior performance in face recognition.  

\subsection{Review of Confidence-Aware Loss Function} 

In \cite{conf/iccv/ShiJ19}, the authors proposed to model face image as a Gaussian distribution, and to use the variation of the face image to define the confidence. This confidence is used as a measure of the face quality.  This approach mitigates the impact of the low quality face images, and reduces the noise effect in training. On top of that, in \cite{journals/corr/abs-2002-11841}, the authors proposed a new loss function, and a few strategies to improve the confidence-aware embedding.   

In this method, each sample $x_i$ is modeled as a Gaussian distribution $N(f_i, \sigma_i^2I)$ in the feature space, where $\sigma_{i}$ is the standard deviation of the feature $f_i$. Define $s_i = 1/\sigma_i^2$, the softmax loss function becomes:

\begin{equation}
    l_4 = -log\frac{e^{s_{i}W_{y_i}^Tf_i+b_{y_i}}}{\sum_{j=1}^n e^{(s_{i}W_{j}^Tf_i+b_{j})}}
    \label{l4}
\end{equation}

Comparing Equ. (\ref{l1}) with Equ. (\ref{l4}), it is obvious that there is an extra term $s_i$ in front of the logit term. The bias terms is simply dropped. After adding a margin term, the loss function is: 

\begin{equation}
    l_{5} = -log\frac{e^{s_iw_{y_i}^Tf_i-m}}{e^{s_iw_{y_i}^Tf_i-m}+\sum_{j\ne{y_i}} e^{s_iw_j^Tf_i}}
    \label{l5}
\end{equation}

Comparing  Equ. (\ref{l5}) with Equ. (\ref{l3}), we see the following two differences. First, Equ. (\ref{l5}) only uses one margin term, while Equ. (\ref{l3}) uses three.  More importantly, there is an extra term $s_i$ in front of the logit term in Equ. (\ref{l5}). This $s_i$ - inverse of the $\sigma_i^2$ - is the quality measure of sample $x_i$.   

We must note that in the loss function Equ. (\ref{l5}), $s_i$ has a indefinite range. Because $m$ is an independent variable, for low quality face images, the value of $s_i$ is hard to go close to 0. As a result, the discrimination capability of $s_i$ is not sufficient. 

\subsection{Proposed Loss Function}

Inspired by the loss function in Equ. (\ref{l3}) and Equ. (\ref{l5}), we proposed a new loss function:

\begin{equation}
    l_{6} = -log\frac{e^{s_iS(m_1\cos(\theta_{y_i}+m_2)-m_3)}}{e^{s_iS(m_1\cos(\theta_{y_i}+m_2)-m_3)} +\sum_{j\ne{y_i}} e^{s_iS\cos(\theta_j)}}
    \label{l6}
\end{equation}
where $s_i$ is in $[0, 1]$, and S has a fixed value, which is equivalent to inverse of the minimum variation value in the Gaussian distribution. In the training process, it is expected that low quality face images should have smaller $s_i$, while high quality face images have larger $s_i$. This $s_i$ makes every sample to use its own quality as its weight in training. Finally, the $s_i$ generated in the model is used as the estimated quality of a face image.  

This loss function combines the advantages of the loss functions in Equ. (\ref{l3}) and Equ. (\ref{l5}).  In Equ. (\ref{l3}), the scalar $s$ is a fixed term, therefore cannot discriminate the quality of face images. However, it does show the best performance in baseline face recognition to our knowledge. The Equ. (\ref{l5}) introduced the term $s_i$ for every sample $x_i$. However since its values is unbounded, it cannot be used in a normalized manner.  In our proposed loss function, both weakness is worked around while the strengths remain.  

\subsection{Proposed Quality Network}

\begin{figure}
    \centering
    \includegraphics[width=8.5cm]{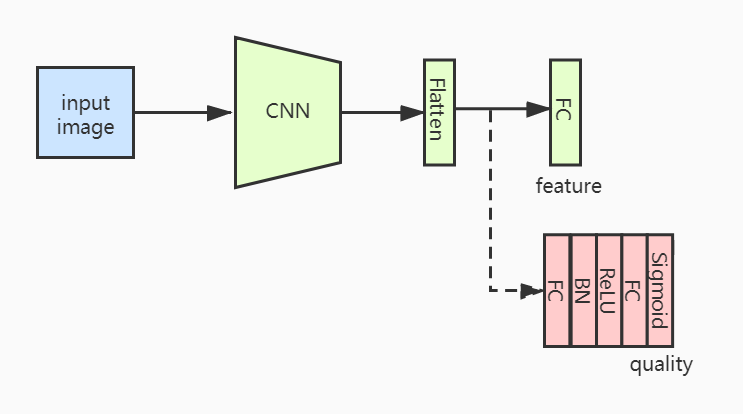}
    \caption{The proposed network architecture.}
    \label{cnn_pic}
\end{figure}

In order for the model to output $s_i$, we add a small quality network as a branch to the baseline network, as shown in Figure~\ref{cnn_pic}. The key layers in this branch network are two FC layers, used to extract the quality value. A BN layer and a ReLU layer are used after the first FC layer, and a sigmoid layer is used after the second FC layer to have a output value in range [0,1]. It is obvious that the complexity of the branch quality network is very small. 

There are three steps in our training pipeline. This training pipeline has some similarity to \cite{conf/iccv/ShiJ19}. We make it an iterative way.   

Step1: $s_i$ is fixed to 1 in the loss function. This is same as the normal face recognition baseline network training, where only the feature extraction is trained.  The face quality network is not trained in this step.

Step 2: After the first step is finished, the baseline network is frozen, and $s_i$ in replaced with the output of the face quality network. The same loss function in Equ. (\ref{l6}) is used for calculation of gradients with regard to the parameters of the face quality network.   

Step 3: The face quality network is frozen and a face quality is generated for every sample face image.  Set $s_i$ to this face quality, and retrain the baseline network. The baseline network can restart from scratch or continue to train from Step 1.   

The Step 2 and Step 3 can iterate more times to train the network to get better performance.  After the Step 3 in first iteration, it is fine to train the backbone and $s_i$ together, but we choose not to do so.  

\begin{figure}
    \centerline{
    \includegraphics[width=6cm]{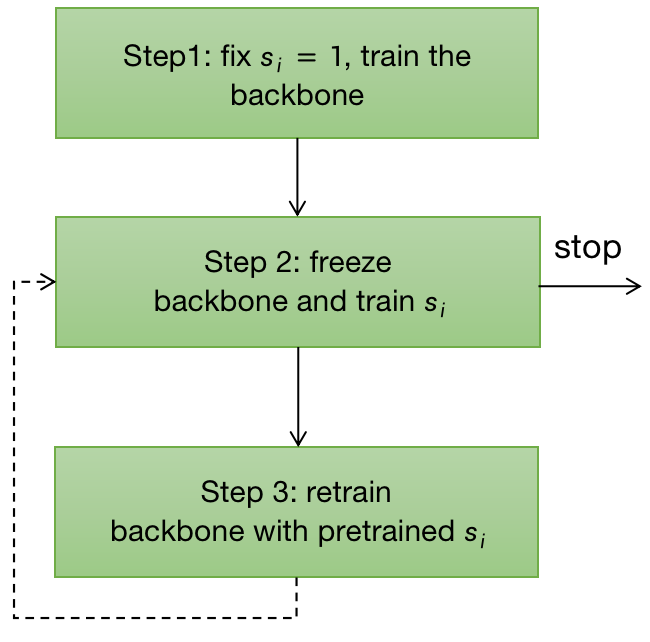}}
    \caption{The training pipeline.}
    \label{pipeline}
\end{figure}

Let us explain why we use this three step training pipeline.  Among multiple face images of a face ID, their qualities vary. Let $w_j$ denote their ground true factor in Figure 3. At the initiation stage of the training process, their angular distance to $w_j$ are large, as shown in Figure 3(a). If we use them directly in the loss function, in order to minimize the loss function, $s_i$ of all face images will be trained to very small value approaching 0. As a result, the quality values of all face images will be small and have no discrimination. Therefore, we need to set $s_i = 1$ at first and only train the baseline network. After this training runs a while, the angular distances of these face images to $w_j$ becomes much smaller. Among them, the angular distances of low quality face images are larger than others. When the training process continues, the $s_i$'s for them will be trained to smaller. As a result, the $s_i$'s are discriminated for high and low quality face images.  One condition for this approach is that there must be more high quality face images than low quality ones in every face identity. In order to ensure their angular distance to $w_j$ is small, in Step 3 we use the face quality value generated in a pre-trained model (Step 2). Since the weights for high quality images are larger, their angular distance to $w_j$ will be trained to even smaller. 

Comparing with \cite{journals/corr/abs-2002-11841} our innovations are three fold. Firstly, we introduce a different loss function using the $s_i$ term and the $S$ term. Please note the subtle difference of the $s_i$ term in the loss functions Equ. (\ref{l6}) and Equ. (\ref{l5}). In Equ. (\ref{l5}), since $s_i$ is in front of $W_{y_i}$ $f_i$ and there is a $-m$ term, $s_i$ will not be trained to 0 for good quality face images. That is why the baseline network and $s_i$ are trained together.  However, in our loss function Equ. (\ref{l6}), the $s_i$ term is outside the $m_1\cos(\theta_{y_i}+m_2)-m_3$ term, so the $s_i$ will be trained to 0 if we start training the baseline network and $s_i$ at the same time. We can change our loss function similar to Equ. (\ref{l5}), but it will cause the same non-normalization issue, since even for very low quality face, the $s_i$ can not be trained to 0. In addition, in the way how our $s_i$ term is defined, we can use the sigmoid activation function to keep $s_i$ in a normalized [0,1] range.  This solves the non-normalization problem in \cite{journals/corr/abs-2002-11841}. With this normalized face quality, we can compare quality of face images of a same person, or of different persons.  

Secondly, we propose a separate small quality network branch to train $s_i$. We can train the $s_i$ in training for every face image, and can generate $s_i$ for every test image in testing. The $s_i$ in \cite{journals/corr/abs-2002-11841} can be trained for the training data, but cannot be generated for test data. Our explicit and quantitative face quality is very important for many applications. 

Thirdly, we propose a very different training pipeline to accomplish the training of $s_i$. In this new training scheme, the qualities of high-quality face images and low-quality face image are more discriminated.   

Lastly, we use the more powerful loss function integrating the margin terms in CosFace, Sphere Face and ArcFace, while only one penalty margin was used in \cite{journals/corr/abs-2002-11841}. It is believed that this loss function is one of the best for face recognition.

\begin{figure}
    \centering
    \includegraphics[width=8.5cm]{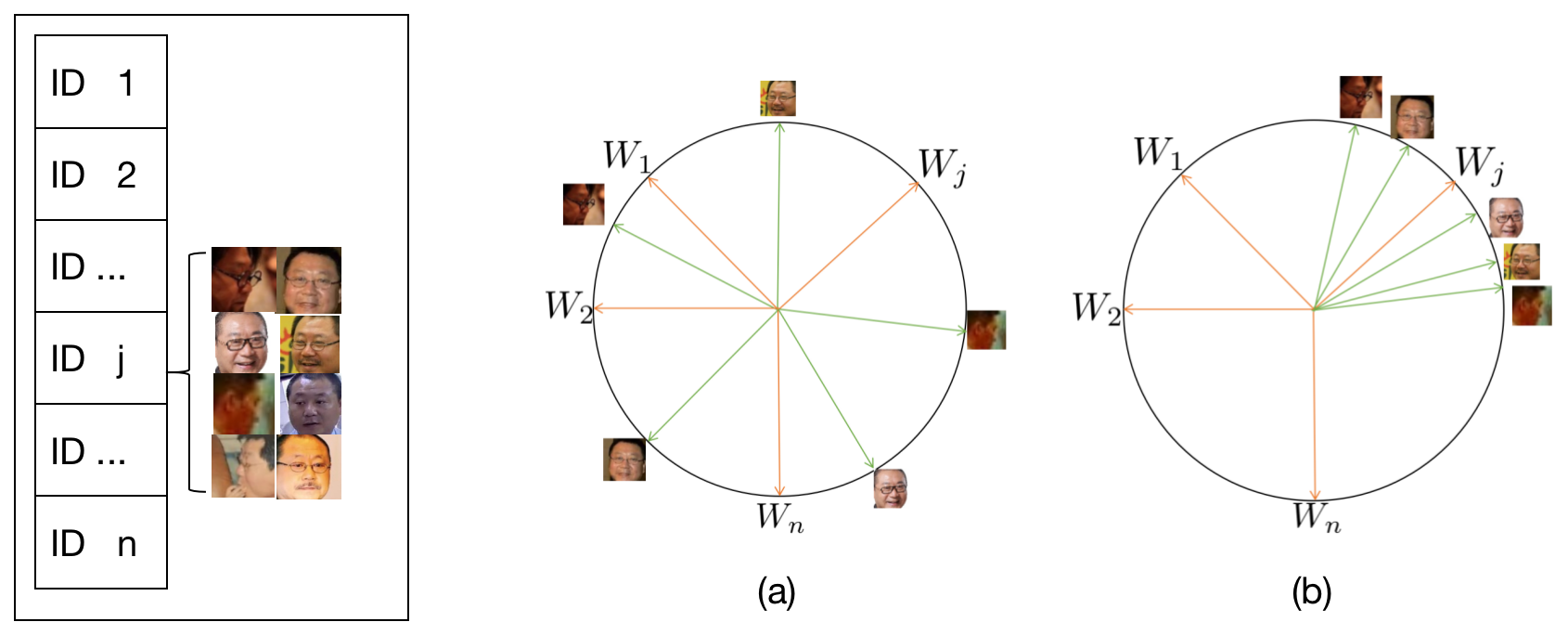}
    \caption{Explanation of the proposed training pipeline.}
    \label{cnn_why}
\end{figure}

%% file: content/experiment.tex
In this section we first describe our implementation details, then describe the datasets we use, and present our experiment results.  

\subsection{Implementation Details}

All the models are implemented with Pytorch v1.6. Our baseline model is essentially a reimplemented ArcFace \cite{ArcFace} and its released code \cite{Insightface}. 

For the training dataset, We use the clean MS-Celeb-1M V2 (MS1MV2) [7] face dataset as in ArcFace [4].  We use the method in \cite{MTCNN} for face alignment and crop all images into a size of 112x112. We use the 100-layer ResNet (R100) as our backbone network. The feature vector size is 512 for all models. We use SGD optimizer with momentum 0.9 and weight decay 5e-4. In the ArcFace loss function, we use $m_1=1.0$, $m_2$=0.3, $m_3$=0.2. We use $S=64$ in our loss function.       
In Step 1 of our training, the learning rate is initially set to 0.1, and decays by 10 at 30, 60, 90 epochs for a total 100 epochs. In Step 2, the learning rate is initially set to 0.01, and decays by 10 at 5, 10 epochs for a total 15 epochs. The learning rate scheme in Step 3 is same as in Step 1. In our benchmark experiments, we stop the training at Step 2 in the 2nd iteration.

\begin{figure*}
    \centering
    \includegraphics[scale=0.275]{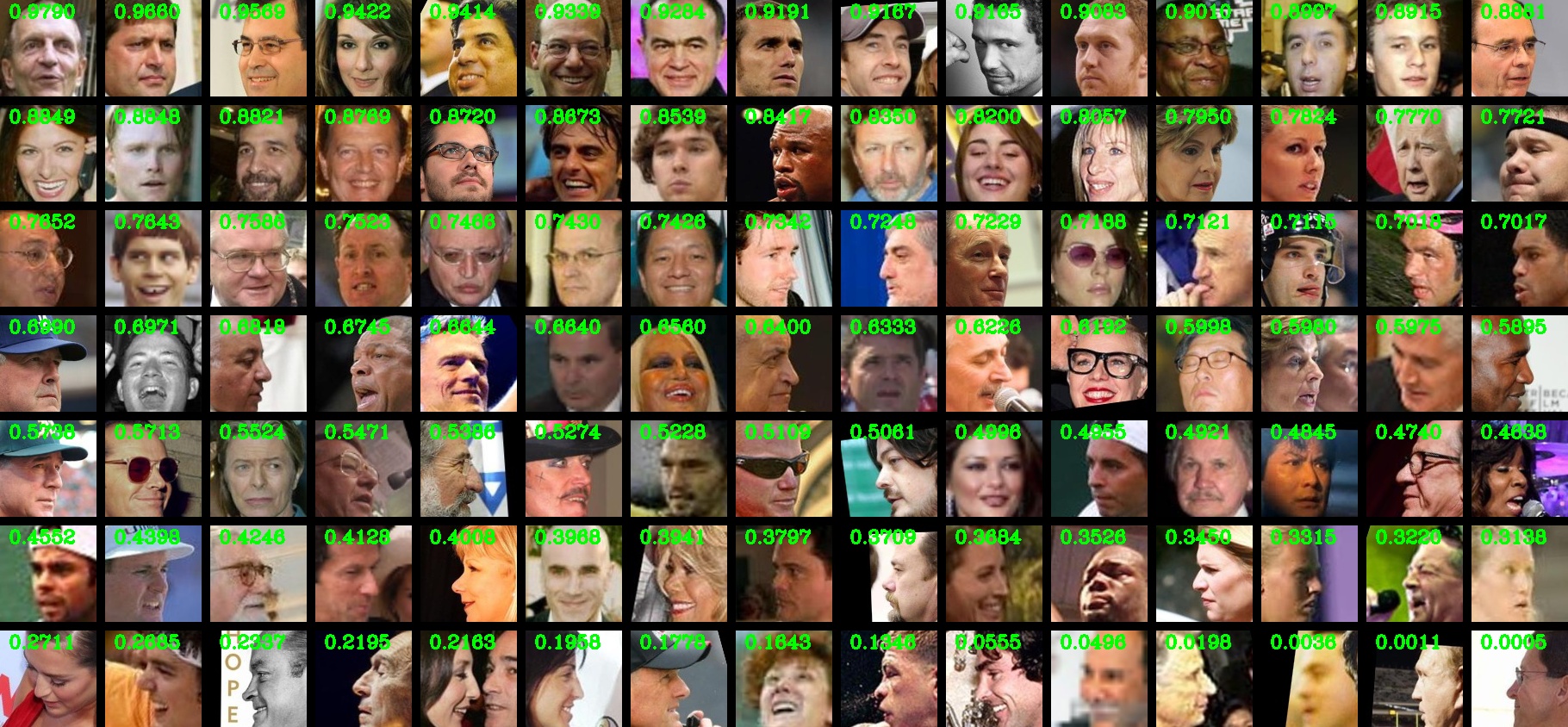}
    \caption{Sample face images and their quality scores}
    \label{qualitysample}
\end{figure*}

After the model training is done, we can generate the quality score for every face image. Unlike other offline methods, like NAN \cite{NAN}, our method can generate a quality score online and for individual face image.  Shown in Figure \ref{qualitysample} are some sample face images from the from the CPLFW \cite{CPLFWTech} dataset and their generated quality scores. We can see that the quality score match the face image very well. 

\begin{table*}[ht]
    \centering
    \begin{tabular}{c|c|c|c|c|c}
        \hline
        Method & LFW\cite{LFWTech} & CALFW\cite{CALFW} & CPLFW\cite{CPLFWTech} &  CFP\_FP\cite{cfp-paper} & YTF  \\
        \hline
        Center Loss\cite{CenterLoss} & 98.75 & 85.48 & 77.48 & - & -\\
        SphereFace\cite{SphereFace} & 99.27 & 90.30 & 81.40 & - & -\\
        VGGFace2\cite{Cao18} & 99.43 & 90.57 & 84.00 & - & -\\
        ArcFace\cite{ArcFace} & \textbf{99.83} & 95.45 & 92.08 & - & 98.02 \\
        Shi et.al \cite{journals/corr/abs-2002-11841} & 99.78 & - & - & \textbf{98.64} & 97.92 \\
        Ranjan et. al \cite{CrystalLoss} & 99.78 & - & - &  - & 96.08 \\
        QAN \cite{QAN} & - & - & - &  - & 96.17+/-0.86 \\
        NAN \cite{NAN} & - & - & - &  - & 95.72+/-0.64 \\
        Liu et. al \cite{FANVFR} & - & - & - &  - & 96.21+/-0.63 \\
        \hline
        \textbf{Ours} - Baseline & 99.80 & 95.91 & 92.55 & 98.34 & - \\
        QW & 99.82 & \textbf{95.98} & \textbf{92.60} & 98.20 & 98.02\\
        QWFA & -  &  -  &  - &  - & 98.12\\
        QWFAF,$s_{th}=0.1$ & -  &  -  &  - &  - & 98.12\\
        QWFAF,$s_{th}=0.2$ & -  &  -  &  - &  - & 98.12\\
        QWFAF,$s_{th}=0.3$ & -  &  -  &  - &  - & \textbf{98.18}\\
        \hline
    \end{tabular}
    \caption{Verification performance (\%) on LFW, CALFW, CPLFW, CFP\_FP, and YTF.}
    \label{compare_sota}
\end{table*}

\subsection{Evaluation Results on LFW, CALFW, CPLFW, CFP-FP, and YTF}

We report the performance of EQFace on the widely used LFW \cite{LFWTech} dataset, and the CALFW \cite{CALFW}, CPLFW \cite{CPLFWTech} and CFP-FP \cite{cfp-paper} datasets. Furthurmore, we also test EQFace on the template video face dataset YTF \cite{YTF}. 

The results are reported in Table 1. In Table 1, the model trained with quality weight (QW) in Equ. (\ref{l6}) is denoted QW, feature aggregation on video template using QW is called QWFA, feature aggregation using QW and a filter on low quality faces is called QWFAF.       

We do four sets of experiments on the YTF dataset.  In the first experiment, we follow the same protocol as in \cite{SphereFace}. Specifically, the feature vectors of the faces in the first 100 frames of a video are averaged. The similarity is calculated on the averaged feature vector between two videos.  This is standard protocol but on a model trained with face quality weight. it is called QW in Table 1.  The second experiment uses the QWFA on first 100 frames.  In the third to fifth experiment, we use the QWFAF with filter threshold $s_{th} = 0.1, 0.2, 0.3$. on the first 100 frames.  If the average quality $mean(s_i)>=s_{th}$ then we weight the feature vectors of qualities $s_i>=s_{th}$. If average quality $mean(s_i)<s_{th}$, then we take the averaged feature vectors same as in the first experiment. 

First of all, our baseline is different from \cite{ArcFace} and \cite{Insightface}.  Even our baseline performance on LFW is worse than \cite{ArcFace}, the QW brings performance improvement on LFW, CALFW and CPLFW. There is a small performance loss on CFP\_FP. The performance improvements on the static dataset shows the implicit benefit of using face quality in the training process.  

The results on the YTF dataset is more convincing.  We notice that the QW accuracy of 0.9802 is same as \cite{ArcFace}. However, when the QWFA is applied, we achieve an accuracy of 0.9812. Furthermore, when we apply extra filter in the QWFAF mode with threshold $s_{th}=0.3$, we achieve an accuracy of 0.9818, the state of art performance on YTF.      

\subsection{Evaluation Results on IJB-B and IJB-C}

In previous section we show that our EQFace surpasses older algorithms \cite{QAN}, \cite{NAN}, \cite{faceQualityPrediction}, \cite{CrystalLoss} by a big margin. So we do not include those algorithms in this section, particularly because they did not have results on the IJB-B and IJB-C datasets.  In this study we use the IJB-B and IJB-C datasets and compare our EQFace with \cite{ArcFace} and \cite{journals/corr/abs-2002-11841}. 

\textbf{The 1:1 verification results} are shown in the 2nd panel in Table 2. The best performances in this panel better than that in the 1st panel are highlighted. Our baseline gives comparable performance to the ArcFace \cite{ArcFace}, which is expected, even though the TAR values vary at different FAR values. On the IJB-C dataset, \cite{journals/corr/abs-2002-11841} achieved the state-of-the-art TAR of 95.0\% and 96.6\% at FAR=1E-5 and 1E-4. The fine tuning method predicts the $\sigma_i$ which has the unbounded issue we explain in Section 3.4. It brings good improvement over the one without fine tuning. The Subcenter-ArcFace \cite{Subcenter} achieves the state-of-the-art performance at most of the FAR values. 

The most important observation is that comparing with the baseline,the QW gives performance gains at all FAR values. The QWFA and QWFAF give noticeable performance gains for FAR at 1E-4 and lower, while give small performance losses for FAR at 1E-3 and 1E-2. Our best TAR performance is better than the Subcenter-ArcFace \cite{Subcenter} at FAR = 1E-6,1E-5 on both the IJB-B and IJB-C datasets, and is very close to \cite{journals/corr/abs-2002-11841} at FAR=1E-5, 1E-4 on the IJB-C dataset.  

Also included in Table 2 are the ablation study results, which will be discussed in more details in next section. 

\begin{table*}
  \centering
  \begin{tabular}{l|l|l|l|l|l|l|l|l|l|l}
    \hline
    \multirow{2}{*}{Method} &
      \multicolumn{5}{c}{IJB-B} &
      \multicolumn{5}{|c}{IJB-C} \\
    & 1E-6 & 1E-5 & 1E-4 & 1E-3 & 1E-2 & 1E-6 & 1E-5 & 1E-4 & 1E-3 & 1E-2\\
    \hline
    ArcFace \cite{ArcFace} & 38.38 & 89.21 & 94.23 & 96.16 & 97.53 & 86.17 & 93.11 & 95.64  & 97.21 & 98.18 \\
    SubCenter-ArcFace \cite{Subcenter} & 35.86 & 91.52 & \textbf{95.13} & \textbf{96.61} & \textbf{97.65} & 90.16 & 94.75 & 96.50 & \textbf{97.61} & \textbf{98.40} \\
    Shi et al. \cite{journals/corr/abs-2002-11841} & - &  - & - &  -  &  - & -  &  91.6 & 93.7 & - & - \\
    Shi et al. \cite{journals/corr/abs-2002-11841} fine tuning & - & - &  - &  -  &  - & -  &  \textbf{95.0} & \textbf{96.6} & - & - \\
    
    \hline
    \textbf{Ours} - Baseline  & 39.30 & 86.74  & 94.27 & 96.41 & 97.61 & 81.97 & 92.01 & 95.62 & 97.36 & 98.27\\
    QW  & 41.90 & 89.62 & 94.51 & 96.48 & 97.62 & 81.41 & 93.01 & 95.84  & 97.39 & 98.30\\
    QWFA  & 46.61 & 91.87 & 94.88 & 96.31 & 97.24 &90.22 & 94.93 & 96.38 & 97.45 & 98.22\\
    QWFAF,$s_{th}$=0.1 & 47.34 & 91.87 & 94.86 & 96.31 & 97.24 & \textbf{90.23} & 94.93 & 96.38 & 97.45 & 98.22\\
    QWFAF,$s_{th}$=0.2 & \textbf{47.30} & \textbf{91.92} & 94.87 & 96.32 & 97.24 & 90.22 & 94.93 & 96.38 & 97.45 & 98.22\\
    QWFAF,$s_{th}$=0.3  & 47.11 & 91.82 & 94.85 & 96.27 & 97.25 & 90.21 & 94.93 & 96.37  & 97.46 & 98.21\\
    \hline
    
    \textbf{Ours} - QWDF,$s_{th}$=0.2   & 41.21 & 89.07 & 94.68  & 96.45 & \textbf{97.77} & 83.22 & 93.67 & 95.96  & 97.38 & \textbf{98.40} \\
    +QWFA, & 45.59 & \textbf{91.99} & 95.05 & 96.20 & 97.52 & 89.72 & \textbf{95.07} & 96.53  & 97.49 & 98.33  \\
    +QWFAF,$s_{th}$=0.1 & 45.60 & 91.99 & 95.05 & 96.20 & 97.52  & 89.73 & 95.06 & 96.53 & 97.49 & 98.33\\
    +QWFAF,$s_{th}$=0.2  & 45.55 & 91.98 & 95.04  & 96.21 & 97.53 & 89.73 & 95.05 & 96.53 & 97.49 & 98.33\\
    +QWFAF,$s_{th}$=0.3  & 44.37 & 91.89 & 95.00  & 96.22 & 97.51 & 89.73 & 95.05 & 96.53 & 97.48 & 98.33\\
    \hline
    
    \textbf{Ours} - QWFA on ArcFace \cite{ArcFace}  & 46.75 & 90.56 & \textbf{94.51} & 95.79 & 97.07 & \textbf{90.15} & \textbf{94.70} & \textbf{96.21}  & 97.17 & 98.05\\
    QWFAF,$s_{th}$=0.1 on ArcFace \cite{ArcFace} & 46.84 & 90.46 & 94.50 & 95.78 & 97.07 & \textbf{90.15} & \textbf{94.70} & 96.19 & 97.18 & 98.05\\
    QWFAF,$s_{th}$=0.2 on ArcFace \cite{ArcFace} & 46.64 & \textbf{90.62} & 94.49  & 95.85 &  97.03 & 90.14 & 94.69 & 96.19 & 97.18 & 98.06 \\
    QWFAF,$s_{th}$=0.3 on ArcFace\cite{ArcFace}  & \textbf{47.53} & 90.47 & 94.45 & 95.80 & 97.01 & 90.14 & 94.69 & 96.18 & 97.19 & 98.06 \\
    \hline
  \end{tabular}\hfill\
  \caption{1:1 verification performance (\%) on IJB-B and IJB-C datasets at FAR=1E-6 to 1E-2. The results of \cite{ArcFace} are regenerated using its code \cite{Insightface}. The results of \cite{Subcenter} are from its Line 14 of Table 2. The 1st and 2nd panels are comparison of our results with the state-of-the-art. The 3rd panel is for ablation study of using quality in distilling training data. The 4th panel is for ablation study of using quality in feature aggregation in testing data, where the highlighted best performance is comparison with ArcFace \cite{ArcFace}.}    
  \label{IJB}
\end{table*}

\textbf{The 1:N identification results} are shown in Table 3. We show our baseline results, which are supposed to be same as ArcFace \cite{ArcFace} but in fact have some varieties in the 4 cases.   

First of all, let us look at the QW results, where QW is used in training, but no feature aggregation is used in testing.  The QW shows performance gain in all four cases by 0.14-0.27\% in the Rank 1 and Rank 5 accuracy. Furthermore, our QW outperforms the ArcFace \cite{ArcFace} and \cite{journals/corr/abs-2002-11841} in the Rank 1 and Rank 5 on the IJB-B dataset, and in the Rank 5 on the IJB-B dataset, as highlighted in the line "QW".  

Next, we test the QWFA and QWFAF. We achieve extra performance gains. The best performance in each case is highlighted. We notice that our best performance in each case is the new state-of-the-art. This proves the benefit of using face quality in training and/or in testing.        

\begin{table}
  \centering
  \begin{tabular}{l|l|l|l|l}
    \hline
    \multirow{ 2}{*}{Method} &
      \multicolumn{2}{c}{IJB-B} &
      \multicolumn{2}{|c}{IJB-C} \\
    & Rank1 & Rank5 & Rank1 & Rank5\\
    \hline
    ArcFace \cite{ArcFace}  & 94.50 & 96.60 & 95.87 & \textbf{97.27}\\
    Shi et.al. \cite{journals/corr/abs-2002-11841}  & - & - &  \textbf{96.00} & 97.06\\
    \hline
    \textbf{Ours} - Baseline  & 94.60 & 96.69 & 95.61 & 97.13\\
    QW  & \textbf{94.79} & \textbf{96.88}  & 95.88 & \textbf{97.27}\\
    QWFA  & 95.03 & 96.60  & \textbf{96.64} & 97.54\\
    QWFAF,$s_{th}$=0.1  & 95.02 & 96.58  & 96.63 & 97.54\\
    QWFAF,$s_{th}$=0.2  & \textbf{95.05} & 96.59  & 96.63 & 97.54\\
    QWFAF,$s_{th}$=0.3  & 95.01 & 96.56  & 96.62 & \textbf{97.57}\\
    \hline
  \end{tabular}\hfill\
  \caption{1:N identification performance (\%) on IJB-B and IJB-C. The ArcFace \cite{ArcFace} results are regenerated using its released code \cite{Insightface}. The Sub-Center ArcFace \cite{Subcenter} results are not included since they are not presented in the paper.} 
  \label{IJB-N}
\end{table}

%% file: content/feature_fuse.tex
The most important ablation study is if and how to use the quantitative quality of a face image in the training and testing of face recognition. These are also the first two applications of them.  That is why we put them in one section.  

\textbf{Other factors in ablation study}: We test multiple iterations of Step 2 and Step 3. We find out that, as long as the Step 3 training converges well in 1st iteration, further iterations of the backbone retraining do not necessary bring extra gains. Our standard configuration is to stop at Step 3 of the 1st iteration. Another possibility is that after the Step 2 of 1st iteration, we can train the backbone and the quality together. But in practice, it is preferred to train them separately to prevent unexpected interaction of the backbone and the quality branch parameters. 

\subsection{Using Face Quality in Training}

In Section 3.4 we describe the training pipeline. We use the face quality metric $s_i$ generated in Step 2 and retrain the baseline network in Step 3. It is in this step where the QW is taking effect. We show in Table 1 and Table 2 the benefits of using the QW in training the face recognition model. 

In Table 1 we show that the extra filter on the low quality faces on the YTF dataset has extra performance improvement.  This motivates us also to explore the filter in the QW in training. In other words, in addition to using $s_i$ as the weight, we force the weight to 0 if $s_i<s_{th}$. Please note that in Table 1 we use the filter on testing data, but in this ablation study we use the same idea on training data. This is one way to distill the training data, and is called quality weighted distilling filter (QWDF). In \cite{Subcenter}, a different approach was used. By studying the distribution of sub-centers, they drop samples whose angular distance is larger than a threshold. 

The results on the IJB-B and IJB-C datasets are shown in the 3rd panel in Table 3. The best performances better than that in the 1st and 2nd panels are highlighted. In this experiment we only test the QWDF with threshold $s_{th}=0.2$ in training. In the testing data, we use QWFAF with $s_{th}=0.1,0.2,0.3$. From the results, we see that we achieve new state-of-the-art performances in a few cases, while in other cases it causes small performance loss comparing with QW without the extra filter in training.        

\subsection{Using Face Quality in Testing}

In Section 4, we show the face quality can be used in the QWFA on video datasets. In this section, we show that the qualify score of the EQFace can be combined with other baseline network, e.g., the ArcFace \cite{ArcFace}. This is equivalent to we use the ArcFace baseline network in Step 1, and stop after Step 2 in the training pipeline. This is similar to our QWFA, and QWFAF, except for there is no QW used in training the baseline network. This is another way how face quality can be used in face recognition testing.  

The results are shown in the 4th panel in Table 2. The best performances are highlighted if they outperform the ArcFace \cite{ArcFace}. We see that for FAR at 1E-4 and lower, the QWFA and QWFAF both have better performance than the ArcFace \cite{ArcFace}. Some of them are better or close to the performance of the Sub-Center ArcFace \cite{Subcenter}.

\subsection{Using Face Quality in Real-Time Video Face Recognition}

In all the performance benchmarks of the video face datasets, we use all the face data throughout a video to aggregate the feature vector. In real-time this is just not possible because not all face images of a person are available at the same time. What is more practical is a progressive face recognition, where a face image is processed online when it shows up in a face recognition device, such as a video camera.     

In this section, we design an experiment to demonstrate how we can use the explicit face quality in a progressive feature aggregation in video face recognition applications. 
Let a person have $N$ face images in a video   
$\{x_1, x_2, \dots,x_N\}$, and their feature vectors and qualities be 
$\{f_1, f_2, \dots, f_N\}$ and $\{s_1, s_2, \dots, s_N\}$. Then at time $i$, the QWFA feature $F_i$ is expressed as:
\begin{equation}
    F_i = \frac{\sum_{j=1}^is_jf_j}{\sum_{j=1}^is_j}
    \label{QWFA}
\end{equation}
A L2 norm is applied afterwards. 

We now introduce the extra filter on the low quality face images, similar to using face quality in testing, but in a progressive manner.  We initially set $F_1=f_1$, $s_{sum}=s_1$, then at time $i$, the QWFA feature $F_i$ is updated as, 

\begin{equation}
    F_i = 
        \begin{cases}
        \frac{s_{sum}F_{i-1}+s_if_i}{s_{sum}+s_i} & if  F_{i-1}\cdot{f_i}>f_{th} \, and \, s_i > s_{th},\\
        F_{i-1} & \, otherwise \,\\
        \end{cases}
    \label{fuse_one_1}
\end{equation}
\begin{equation}
    s_{sum} = 
        \begin{cases}
            s_{sum}+s_i & if  F_{i-1}\cdot{f_i}>f_{th} \, and \, s_i > s_{th},\\
            s_{sum} & \, otherwise \,\\
        \end{cases}
    \label{fuse_one_2}
\end{equation}
where in Equ. (\ref{fuse_one_1}, \ref{fuse_one_2}), $f_{th}$ is the threshold for the similarity of $F_{i-1}$ and $f_i$. The L2 norm is applied on $F_i$.  Please note that, if we ignore the thresholding conditions in these two equations, they simply become the Equ. (\ref{QWFA}). 

In this experiment, we randomly pick 300 IDs and 50 face images for each ID in the vggface2 test dataset and use them as baseline reference dataset. Then we pick 50 more face images for every of these 300 IDs, and these images should be exclusive to the already picked 1500 reference face images. Extra 300 different IDs and 50 images for each ID are picked as a disturbance set. These later 3000 face images are used as the query dataset. 

All the face images in the reference dataset and all images in the query dataset are compared pair-wisely and a 1500x3000 similarity matrix is obtained. We use the known ground truth to calculate the TAR and FAR. In the comparison of the pair-wise face images, the QWFA feature is implemented as in Equ. (\ref{fuse_one_1}, \ref{fuse_one_2}). 

\begin{figure}[t]
    \centering
    \includegraphics[scale=0.35]{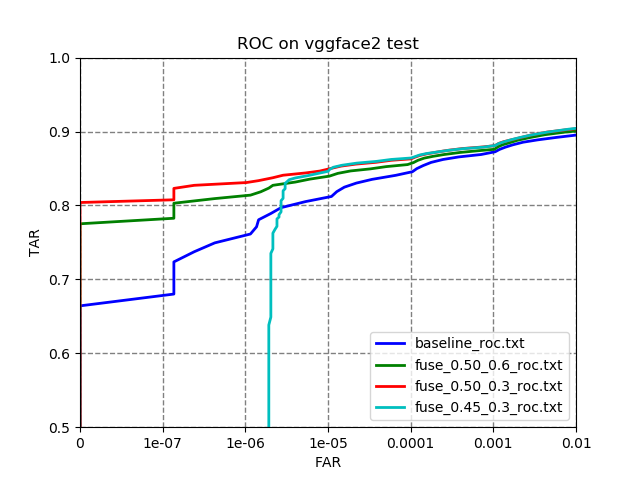}
    \caption{The ROC curve on the vggface2 test data sets. In the legend, 
baseline does not use feature fusion; others use WQFA with threshold $f_{th}$ and $s_{th}$ in Equ. (\ref{fuse_one_1}, \ref{fuse_one_2})}. 
    \label{fuse_roc}
\end{figure}

The performance comparison in the ROC curve is shown in Figure \ref{fuse_roc}.  We can see that, a properly selected face quality threshold can help improve the face recognition accuracy significantly. In contrast, an inappropriate quality threshold may degrade the performance. So the face similarity threshold and the face quality should need to be selected carefully. In our experiment, $f_{th}=0.5$, and $s_{th}=0.3$ gives the best results.

%% file: content/conclusion.tex
In this paper we propose a simple network to estimate the explicit and quantitative quality score for a face image. This explicit quality can be used to distill the training data and to filter low quality testing data. Experiment results prove the performance improvement by the quality weighting in the loss function on a few datasets. As an application, we propose a progressive feature aggregation using this face quality for video face recognition. A designed experiment shows the gain brought by the face quality.

It is worth noting that our approach can be extended to other object or event recognition where the quality of the training or testing image is an important factor.